\documentclass[
]{ceurart}

\usepackage{listings}
\usepackage{booktabs}
\usepackage{graphicx}
\usepackage{multirow}
\begin{document}

\copyrightyear{2026}
\copyrightclause{Copyright for this paper by its authors.
  Use permitted under Creative Commons License Attribution 4.0
  International (CC BY 4.0).}

\conference{Proceedings of the Eighth International Workshop on
  Automated Semantic Analysis of Information in Legal Text
  (ASAIL 2026), 12 June 2026, Singapore}

\title{Re-Ranking Through an Attribution Lens for Citation Quality in Legal QA}

\author[1]{Mohamed Hesham Elganayni}[%
email=mohamed.elganayni@tum.de,
]
\author[1]{Selim Saleh}[%
email=selim.saleh@tum.de,
]

\address[1]{Technical University of Munich, Munich, Germany}

\begin{abstract}
Retrieval-augmented generation systems for legal question answering typically retrieve passages based on semantic similarity and provide them to a language model, which then generates cited answers. Prior work assumes that highly ranked passages are most likely to be usefully cited by the model. Perturbation-based attribution methods, such as C-LIME, have been used exclusively for post-hoc explanation. However, on the AQuAECHR benchmark, semantic similarity does not correlate with passage attribution. Within a retriever's candidate pool, similarity-based ranking performs worse than random selection at surfacing gold citation paragraphs. To address this limitation, a lightweight cross-encoder is trained on continuous perturbation-based attribution scores to re-rank passages prior to generation. This approach is evaluated on the AQuAECHR benchmark, using two language models and five-fold cross-validation. The re-ranker substantially improves citation faithfulness and alignment with gold expert answers. Notably, two re-rankers trained independently on different models converge beyond their raw attribution agreement. This finding indicates that the cross-encoder reduces model-specific noise and produces a shared relevance signal that partially transfers across models, although same-model re-ranking remains more effective. These results demonstrate that perturbation-based attribution provides a practical, model-agnostic training signal for citation-aware retrieval.
\end{abstract}

\begin{keywords}
  retrieval-augmented generation \sep
  legal question answering \sep
  passage re-ranking \sep
  perturbation-based attribution \sep
  citation quality \sep
  cross-encoder \sep
  ECHR
\end{keywords}

\maketitle

\section{Introduction}

Large language models have demonstrated substantial legal reasoning capabilities, including passing bar examinations~\cite{katz2024gpt4bar} and answering jurisdiction-specific questions grounded in case law. However, deployment in legal practice requires more than correct answers. Strict adherence to source text and precise jurisdictional reasoning are essential, and every statement must be traceable to authoritative sources. Retrieval-augmented generation~\cite{lewis2021rag} addresses this by retrieving relevant passages and prompting the model to cite them when generating answers. Nevertheless, citation quality remains a significant bottleneck. For example, on the AQuAECHR benchmark for European Court of Human Rights (ECHR) question answering, retrieve-then-generate systems achieve only modest citation faithfulness~\cite{weidinger2025aquaechr}, and broader evaluations indicate that even the strongest models provide complete citation support for only about half their statements~\cite{gao2023alce}.

A central assumption in these pipelines is that passages ranked highly by the retriever are those the language model will usefully cite. However, this assumption does not hold. On AQuAECHR, semantic similarity between a question and its retrieved passages is effectively uncorrelated with perturbation-based attribution scores, which measure the actual influence of each passage on the generated answer. Furthermore, within the retriever's candidate pool, ranking by similarity performs worse than random selection at surfacing gold citation paragraphs. In specialized domains such as ECHR case law, standard retrievers frequently return topically related but doctrinally irrelevant precedents. Consequently, the generator often ignores the top-ranked evidence in favor of lower-ranked passages or post-rationalizes claims using parametric memory instead of the provided context~\cite{wallat2024correctness}. This highlights that standard retrievers working with embedding-based similarity may be inadequate for identifying the exact paragraphs that a language model actually relies upon for citation.

Perturbation-based attribution methods such as C-LIME and L-SHAP~\cite{monteiropaes2025mexgen} can measure passage-level influence by observing how the generation changes when individual passages are removed. These methods distinguish passages that the model relies on from those it ignores. Despite their effectiveness, they have been used exclusively as post-hoc explanation tools. Cross-encoder re-ranking~\cite{nogueira2020passage} is the established approach for improving passage selection after initial retrieval, but existing re-rankers are trained on relevance judgments or distilled ranking signals rather than on measured generation influence. Yet attribution remains confined to explaining outputs after the fact, and re-ranking remains blind to how the generator actually consumes its input passages.

We address this gap by training a lightweight cross-encoder on continuous C-LIME attribution scores to re-rank passages prior to generation. The approach is evaluated on the full AQuAECHR benchmark using five-fold cross-validation with two language models, Mistral-7B and Llama-3-8B, both used in the AQuAECHR baseline to enable direct comparison. The main contributions are as follows:

\begin{enumerate}
\item We repurpose perturbation-based attribution from a post-hoc explanation tool into a training signal for passage re-ranking, resulting in improvements in citation faithfulness as well as alignment with gold expert answers across both models.
\item We demonstrate that the improvement results from attribution
training rather than the cross-encoder architecture itself. A
pretrained cross-encoder used as a re-ranker without C-LIME training
captures less than half of the citation gains, and its correlation
with attribution scores increases substantially after training on
C-LIME scores.
\item We find that two re-rankers trained independently on different language models converge beyond their raw attribution agreement. This indicates that the cross-encoder reduces model-specific noise and learns a shared passage-relevance signal that transfers across models, achieving consistent improvement over the baseline while retaining 43--51\% of same-model citation gains.
\end{enumerate}

\section{Related Work}

Retrieval-augmented generation~\cite{lewis2021rag} is widely used for producing grounded, cited answers. However, citation quality remains a challenge, as advanced models lack complete citation support for about half of their statements.

A growing body of evidence indicates that this issue results from a fundamental misalignment between retrieval relevance and generation utility. Liu et al.~\cite{liu2023lost} observed that language models display a U-shaped performance curve over retrieved passages, often failing to use relevant information placed in the middle of the input context regardless of its relevance. Randl et al.~\cite{randl2026rage} quantified this gap directly, finding that generators completely disregard the retriever's top-ranked documents for 47--67\% of queries and instead rely on lower-ranked passages 48--66\% of the time. This misalignment also affects citation behavior. Wallat et al.~\cite{wallat2024correctness} demonstrated that up to 57\% of RAG citations are post-rationalized. That is, the model generates content using parametric memory and subsequently selects supporting passages, rather than citing the documents it actually used.

Cross-encoder re-ranking~\cite{nogueira2020passage} is the standard approach for improving passage selection following initial retrieval. The paradigm has since scaled from BERT-based cross-encoders to sequence-to-sequence models such as monoT5~\cite{nogueira2020monoT5} and RankT5~\cite{zhuang2023rankt5}, and more recently to LLM-scale pointwise rerankers like RankLLaMA~\cite{ma2024rankllama} and listwise approaches distilled from proprietary models~\cite{pradeep2023rankzephyr}. Recent literature frames reranking as the critical post-retrieval module for bridging the retriever--generator gap~\cite{gao2024ragsurvey}. Beyond scaling, recent research has begun to adapt re-rankers to align with generator preferences rather than depending solely on surface relevance. Sun et al.~\cite{sun2024rankgpt} showed that LLM ranking judgments can be distilled into lightweight models through permutation distillation, and Jia et al.~\cite{jia2025radio} extended this with RADIO, training re-rankers on LLM-generated text rationales about passage utility. However, these methods depend on the LLM's explicit outputs, which are themselves subject to the faithfulness concerns identified by Wallat et al. Text rationales may indicate what the model claims to find useful rather than what it actually relies on during generation. Notably, none of these approaches examine whether the learned re-ranking signal generalizes across different generator models. Separately, Monteiro Paes et al.~\cite{monteiropaes2025mexgen} introduced C-LIME and L-SHAP, perturbation-based attribution methods that assess each input passage's contribution to a language model's generated output with linear complexity. Unlike rationales or ranking judgments, these scores are derived from observed changes in generation behavior when passages are removed, providing a behavioral measure of passage influence. Nevertheless, C-LIME and L-SHAP have only been designed and evaluated as post-hoc explanation tools, and no prior work has repurposed them as training signals.

Weidinger et al.~\cite{weidinger2025aquaechr} introduced AQuAECHR, a benchmark comprising 1,116 attributed question-answer pairs drawn from European Court of Human Rights case law. Their evaluation, based on the Attributable to Identified Sources framework~\cite{rashkin2022ais} and NLI-based automatic metrics~\cite{gekhman2023trueteacher}, demonstrated that retrieve-then-generate systems achieve 54--70\% citation faithfulness, whereas post-hoc attribution strategies are largely ineffective. Legal retrieval poses distinct challenges: general-purpose retrievers and re-rankers have been shown to hinder performance on legal text due to specialized terminology and fine-grained doctrinal distinctions~\cite{pipitone2024legalbenchrag}, and prior work on ECHR cases has demonstrated that paragraph-level rationale extraction requires domain-specific approaches that account for the argumentative structure of judicial reasoning~\cite{chalkidis2021echr}. This benchmark and evaluation framework serves as the foundation for the present experiments.

Concurrently, Xu et al.~\cite{xu2025scarlet} proposed SCARLet,
which uses perturbation-based utility attribution to train a
bi-encoder retriever via contrastive sampling across
general-domain tasks. Our work differs in three respects: we
train a cross-encoder re-ranker rather than a bi-encoder
retriever, we use continuous attribution scores as regression
targets rather than binary contrastive pairs, and we target
citation quality in legal QA rather than general-domain task
performance. To our knowledge, no prior work has applied
perturbation-based attribution as a training signal for
citation-aware re-ranking, nor investigated whether
attribution-trained re-rankers converge across different
generator models.

\section{Method}

\begin{figure}[t]
\centering
\includegraphics[width=0.85\columnwidth]{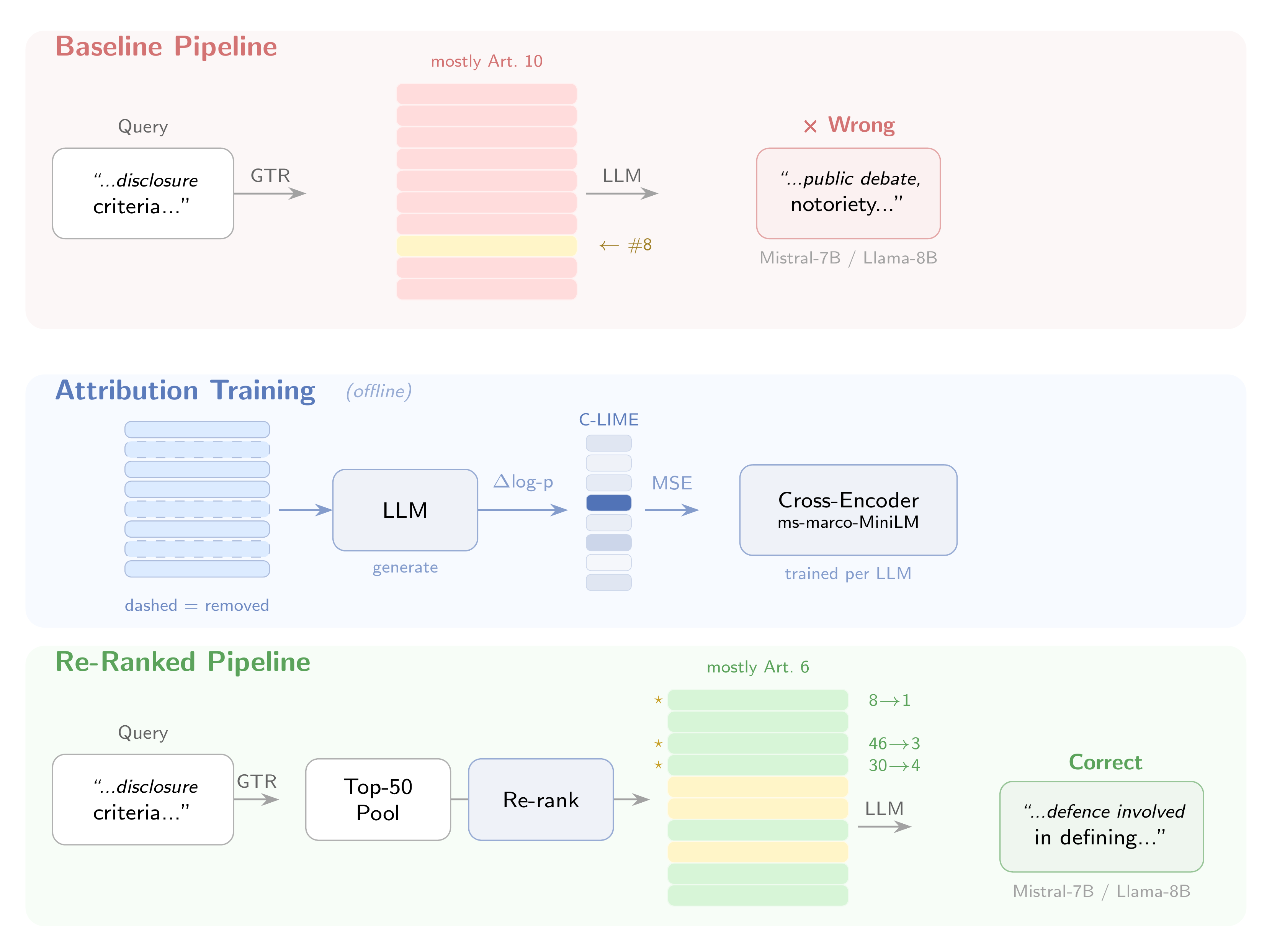}
\caption{C-LIME re-ranking pipeline. Top: baseline retrieval
surfaces passages dominated by Art.~10 media criteria (red).
Middle: perturbation-based attribution trains a cross-encoder
offline. Bottom: re-ranking promotes Art.~6 disclosure passages
(green) to top ranks. Gold-cited passages are marked with both
yellow shading and stars ($\star$). Rank shifts shown at right.}
\label{fig:pipeline}
\end{figure}

Our approach comprises three stages (Figure~\ref{fig:pipeline}). First, we compute perturbation-based attribution scores for retrieved passages. Second, we train a cross-encoder to predict these scores. Third, we use the trained cross-encoder to re-rank passages before answer generation. We evaluate our method on the AQuAECHR benchmark~\cite{weidinger2025aquaechr}, which comprises 1,116 question-answer pairs with gold citations from the European Court of Human Rights. Each question targets a specific legal principle, and the gold answers cite specific paragraphs from ECHR judgments as authoritative sources.
\subsection{Attribution Scoring}

For each question, we retrieve the top 10 passages from AQuAECHR using the GTR-T5-XL dense bi-encoder~\cite{ni2021gtr}, which defines the candidate pool. The language model then generates an answer using greedy decoding. We compute C-LIME attribution scores~\cite{monteiropaes2025mexgen} for all 10 passages at the paragraph level. In C-LIME, each passage is treated as a perturbable unit. The method generates perturbed inputs by removing subsets of passages, observes changes in the generation log probability, and fits a local linear model to estimate each passage's contribution.  We use the log-probability scalarizer, set an oversampling factor of 10, and allow at most two units to be replaced per perturbation. Attribution is computed independently for Mistral-7B-Instruct-v0.2 and Llama-3-8B-Instruct, both in half precision. The resulting scores are continuous and quantify each passage's causal influence on the generated answer, rather than serving as binary citation indicators. 

The key distinction is what each signal measures. A dense retriever scores how closely a passage resembles the query in embedding space, but cannot know how the generator will use that passage. C-LIME instead scores how much a passage changes the generated output when removed. A topically similar but doctrinally irrelevant passage receives a low attribution score because the model's answer is unaffected by its absence, while a passage with low keyword overlap but high doctrinal relevance receives a high score. Training a re-ranker on these scores therefore teaches it to prioritize passages the generator actually consumes.

\subsection{Cross-Encoder Training}

We train a cross-encoder to predict C-LIME scores from question-passage pairs. The architecture is ms-marco-MiniLM-L-6-v2~\cite{reimers2019sbert,nogueira2020passage}, a 22M-parameter model pretrained on relevance judgments. The input is a tokenized [question, passage] pair, truncated to 512 tokens. The output is a scalar score, trained using mean squared error loss with respect to the continuous C-LIME target. The training signal thus encodes how much each passage influenced the generator's output, as measured by C-LIME on the training fold. Continuous targets are preferred because binary cited or uncited labels do not preserve the ranking information necessary for effective passage selection. Training uses AdamW (learning rate $2 \times 10^{-5}$), batch size 32, for five epochs. The 1,116 questions are divided into five folds using stratified question-level partitioning, ensuring that all passages for a given question are assigned to either the training or evaluation set. A separate cross-encoder is trained for each language model and fold, as attribution scores are model-specific, yielding ten cross-encoders in total.

\subsection{Re-Ranking and Generation}

During inference, we use a three-step pipeline for each question. First, we retrieve the top 50 passages using GTR-T5-XL, expanding the candidate pool fivefold compared to the baseline to capture passages that may be missed by similarity ranking. Next, the trained cross-encoder scores all 50 candidates. Finally, we select the 10 top-scoring passages and provide them to the language model using the AQuAECHR baseline prompt. The language model generates an answer using greedy decoding and cites passages with [Doc $i$] markers. To assess cross-model transfer, we also apply the re-ranker trained on one model's attribution scores to re-rank passages for the other model, then generate and evaluate answers in the same way. The re-ranking overhead is minimal. The 22M-parameter cross-encoder scores all 50 candidate passages in approximately 80\,ms on a single A40 GPU, occupying under 440\,MiB of memory. Both the cross-encoder and the 7B-parameter language model fit on the same GPU simultaneously with headroom to spare. No additional LLM calls are introduced, so the re-ranking step amounts to less than 1\% of the total per-question latency, which is dominated by LLM generation.

\subsection{Evaluation}

We use six metrics from AQuAECHR to evaluate both answer correctness and citation quality. For answer correctness, we compute ROUGE-L~\cite{lin2004rouge} to measure lexical overlap with the reference and BERTScore using T5-large~\cite{raffel2020t5} to assess semantic similarity. Claim Recall via TRUETeacher~\cite{gekhman2023trueteacher} evaluates whether each statement in the reference answer is entailed by the full generated answer using an NLI model fine-tuned on T5-11B. For citation quality, Citation Faithfulness measures whether each cited passage supports its corresponding sentence, calculated as the fraction of cited sentences whose concatenated citations entail the statement under TRUETeacher. Exact Match F1 computes the precision and recall of generated citation paragraph identifiers against gold references. NLI-based Citation Similarity assesses whether the generated citations are semantically equivalent to the gold citations by verifying if any generated citation entails each reference block.

\section{Results}

Table~\ref{tab:main} presents the performance of both language models
under baseline and re-ranked conditions across all six evaluation
metrics. Table~\ref{tab:training} reports the cross-encoder's
correlation with C-LIME scores before and after training,
Table~\ref{tab:gold} summarizes gold citation retrieval, and
Table~\ref{tab:transfer} reports cross-model transfer outcomes.

\begin{table}[t]
\centering
\small
\caption{Main results on all 1,116 AQuAECHR questions. R-L = ROUGE-L, BS = BERTScore, EM = Exact Match F1, ClmR = Claim Recall, CitF = Citation Faithfulness, NLI = NLI-based Citation Similarity.}
\label{tab:main}
\begin{tabular}{llcccccc}
\toprule
\textbf{Model} & \textbf{Config.} & \textbf{R-L} & \textbf{BS}
  & \textbf{EM} & \textbf{ClmR} & \textbf{CitF} & \textbf{NLI} \\
\midrule
\multirow{3}{*}{Mistral-7B}
  & Baseline      & 25.61 & 59.25 & 8.86  & 13.27 & 74.76 & 54.65 \\
  & Pretrained RR & 25.65 & 59.41 & 8.95  & 13.35 & 76.44 & 57.35 \\
  & C-LIME RR     & 25.79 & 59.55 & 9.37  & 13.75 & 79.78 & 61.32 \\
\midrule
\multirow{3}{*}{Llama-3-8B}
  & Baseline      & 27.34 & 59.91 & 9.90  & 11.30 & 72.80 & 50.25 \\
  & Pretrained RR & 27.45 & 60.01 & 10.04 & 11.36 & 74.34 & 52.24 \\
  & C-LIME RR     & 27.63 & 60.03 & 10.33 & 11.42 & 75.50 & 55.97 \\
\bottomrule
\end{tabular}
\end{table}

The C-LIME-trained re-ranker consistently improves citation quality
over the baseline across both models, with absolute gains on
Citation Faithfulness of +5.02pp (Mistral) and +2.70pp (Llama), and
on NLI similarity of +6.67pp and +5.72pp, respectively. All six
metrics improve simultaneously for both models. Answer correctness
metrics (ROUGE-L, BERTScore, Claim Recall) change by less than
0.5pp. The pretrained cross-encoder used as a re-ranker without
C-LIME training also improves over the baseline, but captures only
33\% (Mistral) and 57\% (Llama) of the C-LIME CitFaith gain and
40\% and 35\% of the NLI gain, respectively. Mistral exhibits
larger citation gains than Llama across all metrics.

\begin{table}[t]
\centering
\small
\caption{Per-question Spearman correlation between re-ranker
scores and C-LIME attribution (5-fold CV).}
\label{tab:training}
\begin{tabular}{lcc}
\toprule
\textbf{Model} & \textbf{Pretrained $\rho$}
  & \textbf{Trained $\rho$} \\
\midrule
Mistral & 0.187 & 0.686 \\
Llama   & 0.139 & 0.400 \\
\bottomrule
\end{tabular}
\end{table}

A pretrained ms-marco-MiniLM cross-encoder correlates with C-LIME
at Spearman 0.187 (Mistral) and 0.139 (Llama). After training on
C-LIME scores, correlation rises to 0.686 and 0.400, respectively
(Table~\ref{tab:training}).

\begin{table}[t]
\centering
\small
\caption{Gold citation paragraph ranks in the top-50 pool before and after re-ranking. MGR = mean gold rank.}
\label{tab:gold}
\begin{tabular}{lccc}
\toprule
\textbf{Model} & \textbf{GTR MGR} & \textbf{RR MGR} & \textbf{Shift} \\
\midrule
Mistral & 37.4 & 11.1 & +26.3 \\
Llama & 37.4 & 12.0 & +25.4 \\
\bottomrule
\end{tabular}
\end{table}

Under GTR similarity ranking, gold paragraphs fall to a mean rank
of 37.4 out of 50; re-ranking shifts them to 11.1 (Mistral) and
12.0 (Llama).

\begin{table}[t]
\centering
\small
\caption{Cross-model transfer results on all 1,116 questions.
RR = re-ranker. $\rightarrow$ indicates transfer direction.}
\label{tab:transfer}
\begin{tabular}{llcccccc}
\toprule
\textbf{LLM} & \textbf{Re-ranker} & \textbf{R-L} & \textbf{BS}
  & \textbf{EM} & \textbf{ClmR} & \textbf{CitF} & \textbf{NLI} \\
\midrule
\multirow{3}{*}{Mistral}
  & Baseline & 25.61 & 59.25 & 8.86 & 13.27 & 74.76 & 54.65 \\
  & Llama RR $\rightarrow$ Mis. & 25.68 & 59.38 & 9.05 & 13.49
    & 76.92 & 57.83 \\
  & Mistral RR & 25.79 & 59.55 & 9.37 & 13.75 & 79.78 & 61.32 \\
\midrule
\multirow{3}{*}{Llama}
  & Baseline & 27.34 & 59.91 & 9.90 & 11.30 & 72.80 & 50.25 \\
  & Mis. RR $\rightarrow$ Llama & 27.51 & 59.97 & 10.08 & 11.36
    & 74.12 & 53.14 \\
  & Llama RR & 27.63 & 60.03 & 10.33 & 11.42 & 75.50 & 55.97 \\
\bottomrule
\end{tabular}
\end{table}

Both cross-model transfer directions beat the baseline on all six
metrics. The cross-model re-ranker retains 43--49\% of the
same-model CitF gain and 48--51\% of the NLI gain, falling between
the baseline and same-model performance across all metrics.

\section{Discussion}

Our results reveal that the improvement stems from attribution
training rather than cross-encoder architecture. The pretrained
cross-encoder, used as a re-ranker without any C-LIME training,
improves CitFaith by +1.68pp (Mistral) and +1.54pp (Llama) over
the baseline (Table~\ref{tab:main}), confirming that generic
cross-encoder re-ranking helps. However, C-LIME training
roughly triples the Mistral gain (+5.02pp) and nearly doubles
the Llama gain (+2.70pp). This gap is consistent with the
correlation evidence: the pretrained model correlates with C-LIME
at only Spearman 0.187 and 0.139, rising to 0.686 and 0.400 after
training (Table~\ref{tab:training}), indicating that fine-grained
attribution ranking, not the cross-encoder architecture itself,
drives the downstream citation improvements.

The cross-model transfer results reflect a convergence phenomenon.
Raw C-LIME scores between Mistral and Llama agree at only Spearman
0.480, but the two independently trained re-rankers converge to
0.858, indicating that the cross-encoder strips away model-specific
noise. However, this shared signal accounts for only about half the
downstream improvement (43--51\% retention), suggesting that
fine-grained ranking differences still matter for generation quality.

Across both models, NLI gains exceed faithfulness gains (+6.7pp vs
+5.0pp for Mistral; +5.7pp vs +2.7pp for Llama), indicating that
the re-ranker not only improves self-consistency but also shifts
citations toward paragraphs that legal experts selected as
authoritative, despite receiving no direct supervision from gold
citations during training.

The qualitative analysis (Appendix~\ref{app:q1},~\ref{app:q2})
illustrates retrieval failures that aggregate metrics cannot capture.
For an Article~6 disclosure question, the baseline retrieves 8 of
10 passages about Article~10 media criteria due to keyword overlap,
causing the LLM to present the wrong doctrine; the re-ranker
corrects this entirely. For a transgender marriage question, the
re-ranker surfaces the key passage from GTR rank~\#46, absent from
the baseline top-10. Table~\ref{tab:qualitative_ranks} summarizes
the rank changes.

\section{Conclusion}

Similarity-based retrieval fails to surface the passages that legal language models actually cite: on AQuAECHR, the retriever ranks gold citation paragraphs worse than random selection. This work closes that gap by training a lightweight cross-encoder on perturbation-based attribution scores, repurposing a post-hoc explanation tool as a re-ranking signal. The resulting re-ranker improves citation faithfulness by up to +5.0pp and alignment with gold expert citations by up to +6.7pp, without altering answer content. A pretrained cross-encoder without attribution training captures
less than half of these gains, confirming that the improvement
stems from C-LIME training rather than generic re-ranking. Two re-rankers trained on different language models converge to Spearman 0.858 despite raw attribution agreement of only 0.480, confirming that the cross-encoder distills a shared relevance signal; cross-model transfer retains 43--51\% of same-model gains.

For practitioners, a single offline attribution run can produce training data for a re-ranker that improves citation quality at inference time with no additional cost. More broadly, the gap between retriever relevance and generator citability can be narrowed by training on behavioral measures of passage influence rather than on surface relevance or self-reported rationales.

Our evaluation relies on automatic NLI-based metrics, and legal
expert assessment is needed to validate whether the improvements
translate to meaningful gains for practitioners. Because C-LIME
scores reflect generator behavior rather than independent legal
correctness, there is a risk that the re-ranker reinforces
existing model biases rather than promoting objectively relevant
passages. The cross-model convergence finding (Spearman 0.858
despite independent training) partially mitigates this concern,
as two models with different architectures and training data
converge on a shared signal, but domain-expert validation remains
necessary. Extending this approach to larger models, other
jurisdictions, and multilingual settings would test the generality
of the attribution-to-re-ranking paradigm beyond the ECHR domain.

\section*{Declaration on Generative AI}
During the preparation of this work, the author(s) used Claude for coding assistance and \LaTeX{} editing.

\begin{acknowledgments}
This work was funded by the Digitalisierungsinitiative des Bundes
f{\"u}r die Justiz under the ``Generatives Sprachmodell der Justiz
(GSJ)'' project, a collaboration between the justice ministries of
North Rhine-Westphalia and Bavaria, TU Munich, and the University
of Cologne.
\end{acknowledgments}

\bibliography{references}

\appendix

\section{Qualitative Examples}
\label{app:qualitative}

This appendix presents two representative questions contrasting baseline (GTR top-10) and re-ranked (Mistral pointwise) retrieval. The first illustrates a topic-level retrieval failure, where surface keyword matching causes the baseline to retrieve passages from the wrong legal doctrine. The second illustrates deep-pool recovery, where the critical gold passage sits at GTR rank~\#46 and would be unreachable to any top-10 retrieval strategy. Citations originally written as \texttt{[Doc~$i$]} are resolved to case names for readability. \textbf{Bold} marks passages promoted into the top-10 by the re-ranker; \underline{underline} marks passages present in the gold target answer. Table~\ref{tab:qualitative_ranks} summarizes the rank changes.

\begin{table*}[t]
\centering
\small
\caption{Key passage rank changes under re-ranking (Mistral pointwise). GTR = original retriever rank within top-50. RR = re-ranked position in selected top-10. Passages marked with $\star$ appear in the gold citation or target answer. ``--'' indicates the passage was demoted out of the top-10.}
\label{tab:qualitative_ranks}
\begin{tabular}{llccp{6.5cm}}
\toprule
\textbf{Q} & \textbf{Case} & \textbf{GTR} & \textbf{RR} & \textbf{Content Summary} \\
\midrule
\multirow{5}{*}{1}
  & \underline{Sigur{\dh}ur Einarsson v. Iceland \S90} $\star$ & 8 & 1 & Defence involvement in defining disclosure criteria for electronic data \\
  & \textbf{Hurbain v. Belgium \S77} & 23 & 2 & Criteria for data disclosure by protection authorities \\
  & \textbf{\underline{McKeown v. UK \S52}} $\star$ & 30 & 4 & Disclosure judge's role in assessing undisclosed material \\
  & \textbf{Halet v. Luxembourg \S109} & 38 & 9 & Qualifying criteria for public interest disclosures \\
  & \textit{GRA Stiftung v. Switzerland \S56} & 1 & -- & Art.~10 media balancing criteria (irrelevant to Art.~6 disclosure) \\
\midrule
\multirow{4}{*}{2}
  & \textbf{Orlandi v. Italy \S145} & 28 & 1 & Art.~12 applicable to same-sex couples but no obligation to grant marriage \\
  & \textbf{H{\"a}m{\"a}l{\"a}inen v. Finland \S96} & 23 & 2 & Art.~12 as lex specialis; traditional marriage concept \\
  & \textbf{\underline{I. v. UK \S81}} $\star$ & 46 & 3 & Very essence of right to marry infringed for transsexual applicant \\
  & \textbf{Sheffield and Horsham v. UK \S66} & 31 & 7 & Limitations must not impair very essence of right to marry \\
\bottomrule
\end{tabular}
\end{table*}

%% ====== QUESTION 1 ======
\subsection{Question 1: Electronic Data Disclosure}
\label{app:q1}

\noindent\textbf{Question:} \textit{In the context of disclosure of electronic data, what criteria should be considered to determine what might be relevant for disclosure, and why is the defense's involvement in laying down these criteria important?}

\medskip
\noindent\textbf{Target Answer (excerpt):} An important safeguard in the sifting process is to ensure that the defence is provided with an opportunity to be involved in the laying-down of the criteria for determining what might be relevant for disclosure (\underline{\textit{Sigur{\dh}ur Einarsson and Others v. Iceland}}, 2019, \S~90; see also \underline{\textit{Rook v. Germany}}, \S\S~67 and 72).

\medskip
\noindent\textbf{Baseline (Mistral, GTR top-10) --- opening:}

\begin{quote}\small
In the context of disclosure of electronic data, the relevant criteria to determine what might be relevant for disclosure include the contribution to a debate of public interest, the degree of notoriety of the person affected, the subject of the news report, the prior conduct of the person concerned, the content, form and consequences of the publication [\textit{GRA Stiftung v. Switzerland} \S56, \textit{Couderc v. France} \S93, \textit{M.L. and W.W. v. Germany} \S95]. [\ldots] The defense's involvement in laying down these criteria is important because an important safeguard in the process of identifying relevant evidence is to ensure that the defense is provided with an opportunity to be involved in the definition of the criteria for determining what may be relevant [\textit{Sigur{\dh}ur Einarsson v. Iceland} \S90].
\end{quote}

\medskip
\noindent\textbf{Re-ranked (Mistral, pointwise) --- opening:}

\begin{quote}\small
In the context of disclosure of electronic data, the prosecution may be legitimated in sifting through the information to identify what is likely to be relevant [\underline{\textit{Sigur{\dh}ur Einarsson v. Iceland}} \S90]. However, an important safeguard in this process is to ensure that the defense is provided with an opportunity to be involved in the definition of the criteria for determining what may be relevant [\underline{\textit{Sigur{\dh}ur Einarsson v. Iceland}} \S90]. [\ldots] When handling complaints following refusal of delisting by search engines, data protection authorities should consider several criteria to resolve the conflict of rights and interests [\textbf{\textit{Hurbain v. Belgium}} \S77].
\end{quote}

\noindent\textit{Analysis.} Eight of the ten baseline passages concern Article~10 media balancing criteria (public interest, notoriety, prior conduct), matched to the surface keywords ``criteria'' and ``disclosure'' rather than to legal topic, causing the answer to present freedom-of-expression balancing factors as if they were criminal disclosure criteria. The sole relevant passage (\textit{Sigur{\dh}ur Einarsson} \S90) appears only at position 8, and neither \textit{McKeown v. UK} (GTR rank \#30) nor \textit{Rook v. Germany} (GTR rank \#33) --- both cited in the gold answer --- appears in the top-10. With \textit{Sigur{\dh}ur Einarsson} \S90 promoted to position~1, the re-ranked answer opens with the correct legal doctrine, and surfaces \textit{Hurbain v. Belgium} \S77 (from GTR rank \#23) and \textit{McKeown v. UK} \S52 (from GTR rank \#30). Although some Article~10 passages remain in positions 5--8, they no longer dominate the answer, which now correctly addresses Article~6 criminal disclosure rather than Article~10 media balancing.

%% ====== QUESTION 2 (Transgender Marriage) ======
\subsection{Question 2: Transgender Marriage Rights}
\label{app:q2}
\noindent\textbf{Question:} \textit{How does the Court differentiate between the rights of transgender individuals and same-sex couples in relation to the right to marry under Article 12, and what justifies the distinction between a total ban on marriage for transgender individuals and a total ban on same-sex marriage?}

\medskip
\noindent\textbf{Target Answer (excerpt):} Only a total ban on marriage for transgender individuals constitutes a violation of Article 12. The Court found no justification for barring a transsexual from enjoying the right to marry under any circumstances (\underline{\textit{I. v. UK}}, \S81). Conversely, Article 12 does not impose an obligation on Contracting States to grant same-sex couples access to marriage (\textit{Schalk and Kopf v. Austria}, 2010, \S\S~61, 63).

\medskip
\noindent\textbf{Baseline (Mistral, GTR top-10) --- key excerpt:}

\begin{quote}\small
The European Court of Human Rights has differentiated between the rights of transgender individuals and same-sex couples in relation to the right to marry under Article 12 [\textit{Schalk and Kopf v. Austria} \S52]. In earlier cases, the Court found that the attachment to the traditional concept of marriage provided sufficient reason for the continued adoption by the respondent State of biological criteria for determining a person's sex for the purposes of marriage [\textit{Schalk and Kopf} \S51]. [\ldots] In the case of Christine Goodwin, the Court departed from its earlier case-law and found that the impossibility for a post-operative transsexual to marry in her assigned gender violated Article 12 [\textit{Schalk and Kopf} \S52]. [\ldots]
\end{quote}

\medskip
\noindent\textbf{Re-ranked (Mistral, pointwise) --- key excerpt:}

\begin{quote}\small
Article 12 applies to same-sex couples seeking to marry [\textbf{\textit{Orlandi v. Italy}} \S145, \textbf{\textit{H{\"a}m{\"a}l{\"a}inen v. Finland}} \S96]. However, the Court has also recognized that the right to marry under Article 12 is a lex specialis for the traditional concept of marriage [\textbf{\textit{H{\"a}m{\"a}l{\"a}inen v. Finland}} \S96]. [\ldots] The ECtHR has considered whether the allocation of sex in national law to that registered at birth is a limitation impairing the very essence of the right to marry [\textbf{\underline{\textit{I. v. UK}}} \S81]. The Court has found that it is artificial to assert that post-operative transsexuals have not been deprived of the right to marry, as they remain able to marry a person of their former opposite sex [\textbf{\underline{\textit{I. v. UK}}} \S81]. [\ldots] Article 12 cannot be construed as imposing an obligation on the Contracting States to grant access to marriage to same-sex couples [\textbf{\textit{Orlandi v. Italy}} \S145].
\end{quote}

\noindent\textit{Analysis.} The baseline draws almost exclusively from multiple paragraphs of a single case (\textit{Schalk and Kopf} \S\S51--53, 82), because these passages dominate the GTR top-10 through high keyword overlap with the query; the answer traces the doctrinal evolution from \textit{Rees} through \textit{Christine Goodwin} to \textit{Schalk and Kopf}, but lacks \textit{I. v. UK} \S81 (at GTR rank \#46), which contains the ``very essence'' test that justifies treating a total ban on transgender marriage as a violation. After re-ranking, three passages absent from the baseline top-10 anchor the answer: \textit{I. v. UK} \S81 (promoted from GTR rank \#46 to position~3) provides the ``very essence'' test, \textit{Orlandi v. Italy} \S145 (from GTR rank \#28) consolidates the Court's evolving position and extends Article~12's applicability to same-sex couples already married abroad, and \textit{H{\"a}m{\"a}l{\"a}inen v. Finland} \S96 (from GTR rank \#23) characterizes Article~12 as a lex specialis. The result is a structurally clearer answer that separately addresses the transgender and same-sex dimensions with distinct legal standards, rather than cycling through paragraphs of a single judgment.

\end{document}